\pdfoutput=1

\documentclass[11pt]{article}

\usepackage{acl}
\usepackage{microtype}
\usepackage{times}
\usepackage{graphicx}
\usepackage{wrapfig}
\usepackage[inline]{enumitem}
\usepackage{amssymb,multirow}
\usepackage{natbib}
\usepackage{booktabs}
\usepackage{comment}

\title{Transforming Human-Centered AI Collaboration: Redefining Embodied Agents Capabilities through Interactive Grounded Language Instructions}

\author{
Shrestha Mohanty$^1$\thanks{Equal contribution}, 
Negar Arabzadeh$^{2*}$, 
Julia Kiseleva$^1$,
Artem Zholus$^3$,
Milagro Teruel$^4$, \\ 
\textbf{Ahmed Awadallah$^1$,
Yuxuan Sun$^5$,
Kavya Srinet$^5$,
Arthur Szlam$^6$} \\
Microsoft Research$^1$, 
University of Waterloo$^2$, 
École Polytechnique de Montréal$^3$, \\
Universidad Nacional de Córdoba$^4$,
Meta AI$^5$,
DeepMind$^6$ \\
julia.kiseleva@microsoft.com\\
}

\begin{document}
\maketitle
\begin{abstract}
Human intelligence's adaptability is remarkable, allowing us to adjust to new tasks and multi-modal environments swiftly. This skill is evident from a young age as we acquire new abilities and solve problems by imitating others or following natural language instructions.
The research community is actively pursuing the development of interactive "embodied agents" that can engage in natural conversations with humans and assist them with real-world tasks. These agents must possess the ability to promptly request feedback in case communication breaks down or instructions are unclear. Additionally, they must demonstrate proficiency in learning new vocabulary specific to a given domain. 

In this paper, we made the following contributions: 
\begin{enumerate*}[label=(\roman*)]
    \item a crowd-sourcing tool for collecting grounded language instructions;
    \item the largest dataset of grounded language instructions; and
    \item several state-of-the-art baselines.
\end{enumerate*}
These contributions are suitable as a foundation for further research.

\end{abstract}

\section{Introduction}
\label{sec:intro}

Human intelligence possesses the extraordinary ability to adapt rapidly to new tasks and multi-modal environments. This capacity emerges at an early age, as humans acquire new skills and learn to solve problems by imitating others or following natural language instructions. Studies in developmental psychology have proven that natural language communication is a highly effective method of transmitting generic knowledge between individuals, even among infants. This learning approach accelerates the acquisition of new skills by eliminating the need for trial-and-error learning from observations.

One of the long-lasting goals of AI agents~\cite{winograd1972understanding} is the ability to seamlessly interact with humans to assist in solving tasks. To achieve this, the agent must understand and respond to human language to execute instructions in a given environment~\cite{skrynnik2022learning, kiseleva2022interactive,kiseleva2022iglu} or ask clarifying questions~\cite{aliannejadi-etal-2021-building,shi2022learning}, where the end goal is to build an embodied agent. Over the years, researchers have proposed many tasks to tackle this human-AI collaboration challenge, many centered around humans providing instructions to the agent to solve a goal~\cite{gluck2018interactive, shridhar2020alfred}. An early example is the blocks world task, where the agent must understand human instructions to move blocks on a grid~\cite{winograd1972understanding, bisk2016natural}. Other setups use Minecraft~\cite{gray2019craftassist}, such as to move objects around~\cite{abramson2020imitating}, or to simulate human behavior~\cite{park2023generative}.

\begin{figure}
    \centering
    \includegraphics[width=0.5\textwidth]{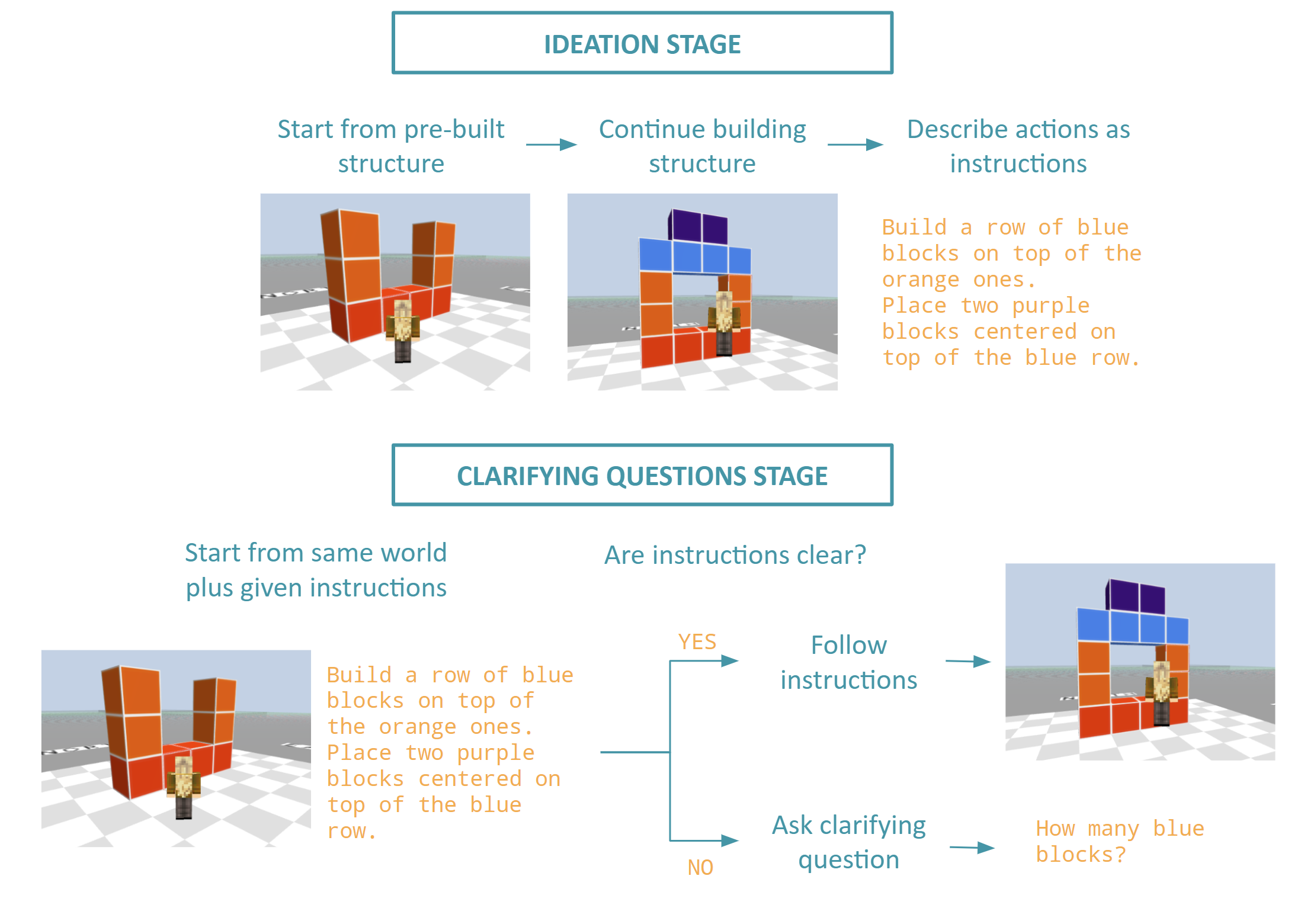}
    \caption{Human-Centered AI Collaboration}
    \label{fig:nlp-task}
\end{figure}

Our paper aims to provide an in-depth investigation into the production of clarifying questions in the context of human-centered AI instruction-based interaction using a Minecraft environment. This scenario presents a unique challenge, as the agent must navigate and complete tasks in a complex, virtual environment, relying solely on natural language instructions. To ensure successful task completion, the agent must accurately identify gaps in the instructions and pose relevant clarifying questions, as demonstrated in Figure~\ref{fig:nlp-task}. 
By tackling this problem head-on, we intend to pave the way for more effective and user-friendly human-AI interactions.

A significant challenge hindering the exploration of clarifying question generation to enhance the user experience during interactions with embodied agents~\cite{narayan-chen-etal-2019-collaborative,bara2021mindcraft} is the scarcity of appropriate datasets and scalable data collection tools. These deficiencies have impeded progress in the field and pose a considerable obstacle to developing effective solutions. Our work addresses this challenge by proposing a novel dataset and scalable data collection methodology, providing a crucial contribution to the field's progress. By addressing this important obstacle, we believe our work will enable researchers to explore new avenues in the field and ultimately enhance user experience in human-AI interactions.

In summary, our main contributions are:
\begin{enumerate}[leftmargin=*,label=\textbf{C\arabic*},nosep]
\item \textbf{Crowdsourcing Tool for Collecting Interactive Grounded Language Instructions:} The development of a crowdsourcing tool specifically designed to efficiently gather interactive grounded language instructions within a Minecraft-like environment at a large scale (Sec.~\ref{sec:data-collection}). This tool facilitates the collection of high-quality data for further research and experimentation.
\item \textbf{Largest Available Dataset of Human-to-Human Grounded Language Instructions:} The creation of an extensive and comprehensive dataset comprising human-to-human grounded language instructions, accompanied by clarifying questions (Sec.~\ref{sec:datasets}). This dataset represents a valuable resource for various research directions, including but not limited to building structures based on given instructions or predicting clarifying questions.
\item \textbf{Baselines for Predicting Clarifying Questions:} The establishment of a set of baselines for the task of predicting clarifying questions that serve as a benchmark for evaluating the performance of future models (Sec.~\ref{sec:baseline}).
\end{enumerate}

\section{Related Work}
\label{sec:rel-work}
Natural Language Interfaces (NLIs) have been a subject of study in various disciplines, including human-computer interaction and information search, for several decades. Early works such as~\cite{woods1972lunar, codd1974seven, hendrix1978developing} laid the foundation for understanding and designing effective interfaces for human language communication with computers.
In recent years, there has been a resurgence of interest in NLIs due to advances in language understanding capabilities driven by large-scale deep learning models~\citep{devlin2018bert, LiuRoberta_2019, clark2020electra, adiwardana2020towards, roller2020recipes, brown2020language, 2303.08774,chowdhery2022palm} and the increasing demand for various applications such as virtual assistants, dialog systems~\cite{li2019dialogue,li2020guided, burtsev2017search,li-etal-2020-rethinking, li2021improving}, semantic parsing, and question answering systems~\citep{liu2017iterative, liu2018adversarial, dinan2020second, zhang2019dialogpt}. The scope of NLIs have expanded from traditional databases to knowledge bases~\citep{copestake1990natural, berant2013semantic} to robots~\citep{tellex2011understanding}, personal assistants~\cite{kiseleva2016understanding, kiseleva2016predicting}, Web service APIs~\citep{su2017building}, and other forms of interaction~\citep{fast2018iris, desai2016program, young2013pomdp}.
The focus has shifted towards interactivity and continuous learning, enabling agents to interact with users, learn new tasks from instructions, assess their uncertainty, ask clarifying questions, seek and leverage human feedback to correct mistakes, and even assess their own mistakes. This includes systems that can learn new tasks from instructions~\citep{li-etal-2020-interactive}, assess their uncertainty~\citep{yao-etal-2019-model}, ask clarifying questions~\citep{Aliannejadi_convAI3, aliannejadi2021building,arabzadeh2022unsupervised}, 
seek and leverage feedback from humans to correct mistakes~\citep{elgohary-etal-2020-speak}, currently LLM can asses their own mistakes~\cite{press2022measuring}.

This paper addresses the important aspect of grounded language understanding, which involves connecting natural language instructions with the real-world context and taking action accordingly. This is crucial to enabling more effective and accurate communication between humans and intelligent agents. Our work focuses specifically on tackling grounded language understanding in the context of collaborative building tasks performed by embodied agents, as highlighted in~\cite{carta2023grounding, kiseleva2021neurips, kiseleva2022iglu,mehta2023improving, mohanty2022collecting,skrynnik2022learning}.

The selection of Minecraft as an environment for grounded language understanding in this work is rooted in a multitude of compelling reasons.
\citet{szlam_why_2019} substantiated the advantages of constructing an open interactive assistant within the sandbox construction game of Minecraft, as opposed to a complex and costly real-world assistant. The Minecraft world's constraints (e.g., coarse 3-d voxel grid and simple physics) and the regularities in the head of the distribution of in-game tasks allow numerous scenarios for grounded NLU research~\citep{yao2020imitation, srinet-etal-2020-craftassist,narayan-chen-etal-2019-collaborative}. Furthermore, the immense popularity of Minecraft as a video game makes it an enticing competition domain, with the second-highest number of total copies sold among all games ever released. This popularity ensures that players will be enthusiastic about interacting with the developed assistants, thus providing a rich resource for human-in-the-loop studies.
Another important advantage of using Minecraft is the availability of the highly developed set of tools for logging agents interactions and deploying agents for evaluation with human-in-the-loop, including: 
\begin{itemize}[nosep, leftmargin=*]
    \item \textit{Malmo}~\citep{johnson2016malmo}: a powerful platform for AI experimentation;
    \item \textit{Craftassist}~\citep{gray_craftassist_2019}: a framework for dialog-enabled interactive agents;
    \item \textit{TaskWorldMod}~\citep{ogawa-etal-2020-gamification}: a platform for situated task-oriented dialog data collection using gamification; and
    \item \textit{MC-Saar-Instruct}~\citep{kohn2020mc}: a platform for Minecraft Instruction Giving Agents;
    \item \textit{IGLU GridWorld}~\cite{zholus2022iglu}: fast environment for training embodied agents.
\end{itemize}

\section{Data Collection Tool}
\label{sec:data-collection}

We developed a tool to enable the collection of multi-modal data such as text, images, and key-value pairs for the collaborative building task~\cite{kiseleva2022interactive,narayan-chen-etal-2019-collaborative, jayannavar2020learning}. This task involves training interactive embodied agents to solve complex tasks while receiving natural language instructions within a collaborative environment.  The interactive agent is defined as follows:
\begin{enumerate*}[label=(\roman*),nosep]
 \item Accurately following instructions in natural language, with a grounding in the current world; 
 \item Seeking clarification when faced with uncertainties;
 \item Swiftly adapting to newly acquired skills.
\end{enumerate*}

For our data collection tool, we strategically harnessed a \emph{Minecraft-like game environment}, which has gained significant popularity and adoption in the NLP and RL communities. Utilizing this environment can overcome the limitations and costs associated with developing and maintaining a real-world assistant~\cite{szlam_why_2019}.
The Minecraft world's unique characteristics, such as its 3D voxel gridworld and adherence to simple rules of physics, provide an abundance of research scenarios and opportunities for experimentation with agents trained by demonstration. The game's interactive nature, including player interaction and dialog exchange, enables grounded language understanding and exploration.
%
Another important factor for consideration is the availability of tools for logging agents' interaction and deploying agents for evaluation with human-in-the-loop within Minecraft. 

\citealp{narayan-chen-etal-2019-collaborative} proposed a setup for \emph{a collaborative building task} with the Minecraft environment where an Architect is provided with a target structure that needs to be built by the Builder. The Architect and Builder communicate with each other through a chat interface. The Architect provides instructions to the Builder on how to create the target structure, and the Builder can ask clarifying questions if an instruction is unclear~\cite{zhang-etal-2021-learning}. The Architect is able to view the actions of the Builder. This approach required installing Microsoft’s Project Malmo~\cite{johnson2016malmo} client, which provides an API for Minecraft agents to chat, build, and the ability to save and load game states. This setup is used to collect multi-turn interactions between the Architect and the Builder collaboratively working towards the common goal of building a given target structure. However, the data collection setup is limited to lab-based studies.

\begin{figure}
    \centering
     \includegraphics[clip, scale=0.35]{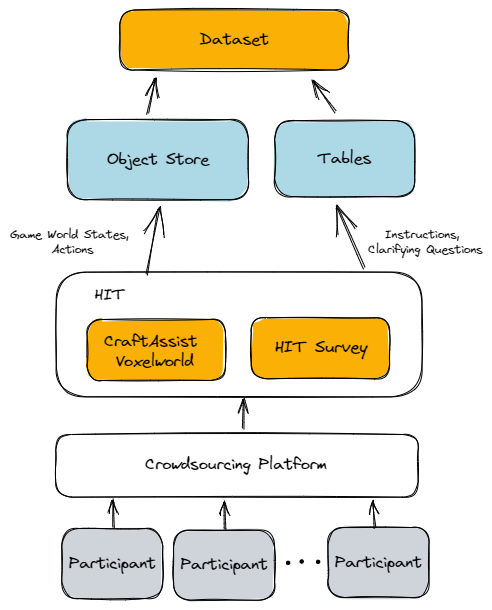}
    \caption{The architecture of the developed data crowdsourcing collection tool}
\label{fig:iglu-data-collection-tool}
\end{figure}

In our work, we have developed and released an open-source data collection tool\footnote{\url{https://github.com/iglu-contest/iglu-data-collection-tool}}. This tool is specifically designed to facilitate the collection of multi-modal collaborative building tasks, seamlessly integrating with crowd-sourcing platforms for efficient participant scaling. Notably, the tool eliminates the need for participants to install a local client, streamlining the data collection process. Figure~\ref{fig:iglu-data-collection-tool} illustrates the overall design of the tool. 

In our study, we have used Amazon Mechanical Turk (MTurk) as the crowd-sourcing platform. Each annotator submits a task referred to as a HIT (Human Intelligence Task). A HIT consists of the CraftAssist~\cite{gray_craftassist_2019} voxelworld along with a HIT survey. The HIT survey is customizable for different tasks and includes rules for a given task, a form where instructions can be submitted, or clarifying questions asked for the building task. CraftAssist is a framework that provides tools and a platform for dialog-enabled interactive agents that learn from natural language interactions. The library provides a 3-d voxelworld grid where agents perform building actions that can be recorded as action states and retrieved for following sessions. Current actions supported by the integrated CraftAssist library include picking, placing, and removing blocks of different colors within the voxelworld. Agents can also jump to place the blocks. These actions enable agents to create structures of varying complexity. Examples of the task or HITs in MTurk along with the embedded voxelworld are provided in the appendix. Finally, the data is stored in two kinds of data stores for ease of access: \emph{Tables} are used to save game ids, instructions, and clarifying questions while the \emph{Object Store} is used for storing files with game world states and collected actions. This data collection tool has been used to collect multi-turn interactions between Architect and Builder similar to the datasets collected by ~\citealp{narayan-chen-etal-2019-collaborative} described next. 
\section{Datasets}
\label{sec:datasets}

We used the previously described data collection tool to build corpora of multi-modal data which could be used towards solving wide-ranging NLP and RL tasks including training interactive agents by demonstrations given natural language instructions~\cite{skrynnik2022learning}. 
Our research initially concentrates on multi-turn interactions, following a similar approach as presented by ~\cite{narayan-chen-etal-2019-collaborative} (Sec.~\ref{sec:multi-turn-dataset}). To enhance the size of our dataset, we subsequently expanded our data collection efforts to a Single-Turn dataset (Sec.~\ref{sec:single-turn-dataset}). This approach allowed us to gather a larger corpus of data more efficiently.
The datasets and accompanying code for analysis and visualization have been made openly available~\footnote{\url{https://github.com/microsoft/iglu-datasets}}.

\subsection{Multi-Turn Dataset}
\label{sec:multi-turn-dataset}

\begin{figure}
    \centering
     \includegraphics[clip, scale=0.35]{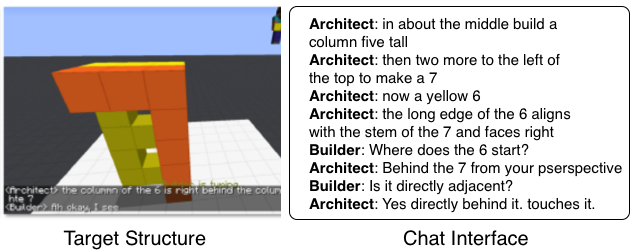}
    \caption{The example of the game from the multi-turn dataset, where  Architect can see the target structure and needs to provide instructions for the Builder~\cite{mehta2023improving}}
\label{fig:multi-turn}
\end{figure}

The Multi-Turn dataset comprises dialog-behavior sequences, which we called \emph{game}, as depicted in Figure~\ref{fig:multi-turn}. In each \emph{turn}, an annotator takes on the role of either the Architect or the Builder. The Architect provides the next step instruction, while the Builder executes the instruction or poses a clarifying question. The sequences follow a linear progression without branching out, starting from scratch for a given goal structure or building on top of intermediate results. The goal structures used in the dataset are sourced from~\cite{narayan-chen-etal-2019-collaborative}. 

\begin{table}[]
\caption{Statistics of Multi-Turn Dataset}
\begin{tabular}{@{}ll@{}}
\toprule
Target Structures                                                             & 31          \\
Completed Games                                                               & 47          \\
\begin{tabular}[c]{@{}l@{}}Median Duration of \\ Completed Games\end{tabular} & 59 mins     \\
Utterances                                                                    & 871         \\
Avg. Length of Instructions                                                   & 19.32 words \\
Clarifying Questions                                                          & 126         \\ \bottomrule
\end{tabular}
\label{tab:multiturn_data}
\end{table}

Tab.~\ref{tab:multiturn_data} shows the summary of the Multi-Turn dataset. There are 31 goal structures presented to annotators to build. We process and clean the data by filtering out missing and low-quality submissions such as very short instructions. Finally, we have 47 completed game sessions with the median duration of a game being around 1 hour. A game session is considered complete when the Builder is able to create a given goal structure after interacting with and following instructions provided by the Architect. This is denoted by the Architect marking the structure as \emph{``complete"}. Across all the games, there were 871 number of utterances or dialog interactions between the Architect and Builder annotators. The average length of instructions provided by the Architects was around 19 words, and the number of clarifying questions asked by the Builders -- 126.

To provide a deeper understanding of the covered structures in our multi-turn dataset, we performed manual labeling on the 31 structures. The labels, along with their meanings and the corresponding number of structures in the dataset in brackets, are as follows:

\begin{enumerate}[label=(\roman*),leftmargin=*, nosep]
    \item \textit{flat [7]:} all blocks on the ground
    \item \textit{flying [27]:} there are blocks that cannot be fully-added without removing some other blocks 
    \item \textit{diagonal [6]:} some blocks are adjacent (in the vertical axis) diagonally
    \item \textit{tricky [6]:} some blocks are hidden or there should be a specific order in which they should be placed
    \item \textit{tall [25]:} a structure cannot be built without the agent being high enough (the placement radius is 3 blocks)
\end{enumerate}

\subsection{Single-Turn Dataset}
\label{sec:single-turn-dataset}

From our extensive study on Multi-Turn data collection, we identified certain challenges that crowdsource workers encountered when engaging in the collaborative building task and issuing instructions for specific target structures. To streamline and enhance the crowd-sourcing process, we decided to simplify the task.
 
Our approach involved removing the added complexity of building a predefined target structure. Instead, participants were free to perform free-form building actions within the voxelworld while providing instructions  that should allow another worker to rebuild the same structure.
This modification led to creating Single-Turn task segments, where participants collaborated asynchronously to construct the same structure. This adjustment enabled us to collect data at a faster pace, resulting in a larger corpus comprising natural language instructions, corresponding actions performed based on those instructions, and a set of clarifying questions. We record and save actions performed by annotators in a key-value pair format that stores the movement of the agent and positional changes of blocks within the voxelworld.

To provide diverse starting canvases for annotators, we utilized the Multi-Turn dataset to load different world states, which served as varying initial conditions for the building process. The process of collecting single-turn instructions and associated clarifying questions is illustrated in Figure~\ref{fig:nlp-task}. The detailed procedure is outlined below:
\begin{itemize}[leftmargin=*, nosep]
    \item An annotator is assigned a world state from the Multi-Turn dataset as the starting point for their building task (Figure~\ref{fig:nlp-task}: Ideation Stage).
    \item The annotator is prompted to perform a sequence of actions for a duration of one minute.
    \item Then, the annotator is required to describe their set of actions in the form of instruction.
    \item Another annotator is shown the instruction and is asked to perform the steps mentioned. If the instruction is unclear, the annotator specifies it as thus and asks clarification questions (Figure~\ref{fig:nlp-task}: Clarification Question Stage).
\end{itemize}

Tab.~\ref{tab:single-turn-stat} presents comprehensive statistics on the Single-Turn dataset, currently the largest dataset available for interactive grounded language understanding. We processed and cleaned the collected Single-Turn dataset by following a heuristic approach which included filtering out samples where the length of instruction was very short. We also checked whether the instruction was in English and manually evaluated jobs to remove submissions by annotators who provided low-quality instructions such as providing the same instruction repeatedly. As shown in Table~\ref{tab:single-turn-stat}, the Single-Turn corpus consists of 8,136 pairs of actions and instructions. On average, an instruction has 18 words which are indicative of the instructions being descriptive enough for a one-minute building process.

In addition to the processing steps for cleaning instructions, for the clarifying questions we furthermore verified if the annotator marked the instruction as ambiguous, they must have issued a clarifying question else the submission would be filtered out with a warning provided to the annotator. This was to ensure that every instruction annotated as ``not clear" is accompanied by at least one clarifying question. Out of 8,136 instructions, 1,056 (12.98\%) were annotated as \textit{Not Clear} thus being ambiguous, and 7,080 (87.02\%) as \textit{Clear} instructions. The average length of clarifying questions is around 12 words.

Tab.~\ref{tab:examples-single-turn} exemplifies a few instructions marked as being unclear along with clarifying questions issued by annotators. The majority of clarifying questions fall into the categories below:
\begin{itemize}[leftmargin=*, nosep]
\item \textit{Color of blocks}: Questions clarifying the color of the blocks to be used. For instance, an instruction specified \emph{``Place four blocks to the east of the highest block horizontally."} The corresponding clarifying question issued was \emph{``Which color blocks?"}
\item \textit{Direction or Orientation}: Questions that clarify the direction and orientation to build in the world. For example, given the instruction \emph{``break three blue blocks and place two green ones."} The clarifying question issued was \emph{``Where do I place two green ones?"}
\item \textit{Number of blocks}: Questions that clarify the number of blocks to be placed. For example, given the instruction \emph{``Built yellow blocks across the north."} The clarifying question issued was \emph{``How many yellow blocks should be built?"}
\item \textit{Identifying blocks to be changed}: Questions posed to identify specific blocks that need to be changed.
For instance, given the instruction \emph{``Destroy the 3 stacked red blocks on the east side. Replace them with 3 stacked blue boxes."} The clarifying question issued was \emph{``Which three of the four stacked red blocks on the east side need to be destroyed?"}
\end{itemize}

The Single-Turn approach offers several advantages over the sequential nature of the Multi-Turn process. One significant advantage is the independence of each sample, which allows for easier utilization in different tasks. Each turn can be interpreted as a complete set of information, enabling flexibility and versatility in its application.
In the Single-Turn approach, samples can be easily plugged into different settings or scenarios, as they do not rely on the context of previous turns. This independence allows researchers to extract valuable insights and information from individual turns without the need to consider the entire dialogue sequence.
Furthermore, the Single-Turn approach allows for the collection of multiple clarifying questions for each instruction. This enhances the richness and diversity of the dataset, enabling a deeper understanding of the nuances and challenges in generating clarifying questions.

\begin{table}[]
\caption{Statistics of Single-Turn Dataset}
\begin{tabular}{@{}llll@{}}
\toprule
\multicolumn{2}{l}{Instructions} & \multicolumn{2}{l}{Avg. Length (in words)} \\ \midrule
Number           & 8136         & Instructions                     & 18.29       \\
Clear            & 7080         & Clarifying  Questions        & 12.05       \\
Ambiguous        & 1056         &                                &             \\ \bottomrule
\end{tabular}
\label{tab:single-turn-stat}
\end{table}

\begin{table}[!t]
\centering
\caption{Examples of pairs of unclear instructions and clarifying questions}
\scalebox{0.5}{
\begin{tabular}{p{10.5cm}|p{5.5cm}}

\\\textbf{Unclear Instruction} & \textbf{Clarifying Question}
\\[0.5em] 
\hline
Place four blocks to the east of the highest block, horizontally. & Which color blocks?\\[0.5em] \hline
Destroy 2 purple blocks and then build 3   green blocks diagonally. & Which two purple   blocks need to be destroyed? \\ \hline
Destroy the 3 stacked red blocks on the east side. Replace them with 3 stacked blue boxes & Which three of the four stacked red   blocks on the east side need to be destroyed? \\ \hline
Make a rectangle that is the width and height of the blue shape and fill it in with purple blocks. & Which side I need to make the rectangle is not clear \\ \hline
Facing South remove the rightmost purple block. Place a row of three   orange blocks to the left of the upper leftmost purple block. Place two   orange blocks above and below the leftmost orange block. & Which one of the rightmost blocks should be removed? \\ \hline
Facing north and purple green blocks will arrange one by one. & Where would you like to place the purple and green blocks exactly? \\ \hline
Built yellow blocks across the north. & How many yellow blocks should be built? \\ 
\end{tabular}}
\label{tab:examples-single-turn}
\end{table}

\section{Baselines Models and Evaluation}
\label{sec:baseline}
The collected dataset (Sec.~\ref{sec:single-turn-dataset}) offers an opportunity to delve deeply into the exploration of the following key research questions:
\begin{itemize}[leftmargin=*, nosep]
    \item \textbf {When to ask clarifying questions?}: This research question aims to predict whether an instruction provided by the Architect is sufficient for the Builder to complete a task successfully or if further clarification is required. 
    \item \textbf{What clarifying question to ask?} 
    When faced with an instruction that is considered ambiguous, this research question focuses on determining the appropriate question to ask for further clarification. 
\end{itemize}
\noindent
It is worth noting that issues related to determining ``When" and ``What" to ask as clarifying questions have gained significant attention in the domains of Natural Language Processing and Information Retrieval \cite{aliannejadi2019asking,aliannejadi2021building,aliannejadi2020convai3,arabzadeh2022unsupervised}. However, as far as our knowledge goes, this aspect has not been explored to a great extent in the context of interacting with intelligent embodied agents. In the following sections, we present two end-to-end pipelines that have shown promising performance in addressing each research question.

\subsection{When: Clarification Need Prediction}
\label{sec:clarification-need}
Al the baselines publicly are available.\footnote{\url{https://github.com/iglu-contest/nlp-baselines-2022}} All results are reported in Table~\ref{tab:task1res}. In line with the nature of the task, we utilize the F-1 Score as the evaluation metric since it provides a balanced measure of precision and recall, offering valuable insights into the performance of the classification model.

\begin{table}[]
\caption{Results of the baselines on `When': Clarification Need Prediction task}
\centering
\scalebox{0.8}{
\begin{tabular}{ll}
\hline
Baseline & F-1 score \\ \hline
Fine-tuned BERT (Sec.~\ref{sec:baseline-bert}) & 0.732 \\
Text-Grid Cross Modularity (Sec.~\ref{sec:baseline3}) & 0.757 \\
Textual Grid world State (Sec.~\ref{sec:text-grid-baseline}) & 0.761 \\ 
 \hline
\end{tabular}}
\label{tab:task1res}
\end{table}

\subsubsection{BERT fine-tuning} 
\label{sec:baseline-bert}

Our dataset provides a substantial amount of training data. Therefore, as suggested in~\cite{aliannejadi2021building}, the simplest baseline to determine whether an instruction requires a clarifying question would be fine-tuning LLMs such as BERT~\cite{devlin2018bert}. This is followed by a classification layer to predict whether the instructions are clear. This approach has shown promising performance on similar classification tasks~\cite{aliannejadi2021building,arabzadeh2022unsupervised} demonstrated in Tab.~\ref{tab:task1res}. 

\subsubsection{Text-Grid Cross Modularity} 
\label{sec:baseline3}
This baseline \cite{shi2023ask}, which has shown improved performance compared to the simple LLM fine-tuning approach, consists of the following four major components: 
\begin{enumerate} [leftmargin=*,label=\textbf{(\roman*)}, nosep]
    \item \emph{Utterance Encoder}, where Architect and Builder annotations would be added before each architect utterance $A_t$ and each builder utterance $B_t$, respectively. Then, the dialogue utterances are represented as $D_t = ``architect'' A_t \oplus ``builder'' B_t$ at the turn $t$, where $\oplus$ is the operation of sequence concatenation. The dialogue is encoded through pre-trained language models such as BERT.
    \item \emph{World state encoder}which aims to represent the pre-built structure using a voxel-based grid. Each grid state is encoded as a 7-dimensional one-hot vector, representing either an empty space or a block of one of six colors. This encoding results in a $7 \times 11 \times 9 \times 11$ representation of the world state.
    The structure of the World State Encoder is similar to the approach presented in \cite{jayannavar2020learning}. It comprises k 3D-convolutional layers followed by a Rectified Linear Unit (ReLU) activation function. This configuration allows the encoder to extract meaningful features from the voxel-based grid representation of the world state. By applying convolutional layers and non-linear activation, the World State Encoder captures spatial dependencies and abstract representations of the pre-built structure. 
    \item \emph{Fusion module} which consists of three sub-modules: one Single-Modality and two Cross-Modality. The former modules are based on self-attention layers and the latter on cross-attention layers. These take as input the world state representation and dialogue history representation. Between every successive pair of grid single-modality modules or text single-modality modules, there is a cross-modality module. 
    \item  \emph{Slot Decoder}, this component contains one linear projection to obtain a scalar value for the final binary classification through the Sigmoid function.  
\end{enumerate}

\subsubsection{Textual Grid world State}
\label{sec:text-grid-baseline}
This baseline focuses on mapping the GridWorld state to a textual context, which is then added as a prefix to the verbalizations of the Architect-Agent. This approach has demonstrated its effectiveness in the classification task by utilizing an automated description of the number of blocks per color in the pre-built structures. This additional information conveyed through the textual description, has proven to be valuable for the classification task.

For instance, a voxel world can be automatically converted into a textual description like: \emph{``There are 4 levels. There are 15 different blocks. At level 0, there are 3 green blocks. Above the 1st level, there are 2 purple, 2 yellow, and 1 green block. Above at level 2, there are 3 green blocks. Above the 3rd level, there are 2 yellow and 2 green blocks.''} This description provides important contextual information about the voxel world and contributes to the improved performance of the simple LLM fine-tuning baseline.
Overall, the inclusion of a textual description of the voxel world has enhanced the simple LLM fine-tuning baseline by 4\% in terms of performance (Tab.~\ref{tab:task1res}. This approach showcases the importance of incorporating relevant contextual information to enhance the understanding and classification of language-guided collaborative building tasks.

\subsection{What: Clarifying Question Retrieval}
\label{sec:what}
We formulate it as ranking a pool of clarifying questions based on their relevance to ambiguous instructions to place the most pertinent clarifying questions at the top of the ranked list. Given that the relevance judgments for this task are sparse. Namely, only one clarifying question per ambiguous instruction is annotated. We evaluate the task using the Mean Reciprocal Rank (MRR) at cut-off 20. This evaluation approach is consistent with well-known benchmarks like MS MARCO~\cite{nguyen2016ms}.

Tab.~\ref{tab:task2res} presents the performance of BM25 followed by the two introduced baselines, measured using the MRR@20.

\begin{table}[]
\caption{Results of the baselines on `What': Clarification Need Prediction task}
\centering
\scalebox{0.8}{
\begin{tabular}{ll}
Baseline & MRR@20\\ \hline
BM25 & 0.3410	\\ 
Baseline 1 (Sec.~\ref{sec:what-baseline1}) & 0.5360	 \\
Baseline 2 (Sec.~\ref{sec:what-baseline2}) & 0.5960	\\ \hline
\end{tabular}}
\label{tab:task2res}
\end{table}

\subsubsection{Baseline 1}
\label{sec:what-baseline1}
\emph{Text Representation:} A frozen DeBERTa-v3-base model has demonstrated promising performance for ranking tasks. The instructions are encoded in this baseline, followed by a separator and a question. The last four layers of DeBERTa are concatenated and passed through a two-layer BiLSTM to acquire a text representation.

\emph{World Representation:} A world representation is utilized to create a 3D grid. This is subsequently passed through a 1D convolutional network to simplify the height dimension (y), and then the resulting vector is passed through a 2D convolutional network to reduce the width/length (x, z) dimensions. The underlying assumption is that height occupies a different semantic space from the interchangeable x, z dimensions. For example, an instruction might include references to a ``tower" or ``column," which would be a stack of objects in the y direction, while a ``wall" could extend in the x or z direction. Ultimately, the size of the 3D grid is reduced by an AvgPooling layer to a 1D vector.

Subsequently, the encoded text representation and the world representation are concatenated, and the vector is passed through a two-layer MLP to obtain the final representation. The model is trained using a CrossEntropy loss function over 10 folds cross-validation. At inference time, the ensemble predictions of the 10 models are used for the final predictions.

In addition, it has been revealed that certain straightforward post-processing tricks can enhance performance. These post-processing methods rely on certain assumptions about the content of questions given a world and an instruction. For example, the size of the ranking pool could be reduced by excluding questions that don't overlap with the given instructions. If the instruction doesn't mention a color like ``blue," and ``blue" is also absent in the world, it can be assumed that the question won't reference the word ``blue." While these heuristic rules may seem somewhat aggressive, they have proven useful in excluding additional questions irrelevant to the instruction.

\subsubsection{Baseline 2}
\label{sec:what-baseline2}

To comprehend the concept of relevance, the approach of aligning queries and relevant items closely in embedding space while distancing queries from irrelevant items in the same space has proven to be effective~\cite{izacard2021towards,reimers2019sentence,karpukhin2020dense,zhan2021optimizing}. Similarly, in this baseline, each positive question is paired with sampled irrelevant negative questions drawn from the candidate questions. The similarity between the instruction and the question is then measured using a BERT-like pre-trained language model.

To include information from the world state and pre-built structure, it is recommended to encode state information, such as the colors and numbers of initialized blocks, in the form of natural language and then concatenate this with the instruction. It has been demonstrated that clarifying questions about the same instruction can differ based on the world states. Therefore, to avoid redundant state information and improve the model's robustness and generalization, randomly selecting only one color type of block as the state information has proven helpful and has increased the model's generalizability. The state information and raw instruction are then concatenated and labeled with the keywords ``state" and ``instruction," respectively. For instance, the input could be: ``state: There are nine green blocks; instruction: ``put a green block on top of the yellow and the two blue ones.''

Before moving on to the training phase, and to balance the data distribution through augmentation, Easy Data Augmentation (EDA) has been shown to be effective~\cite{DBLP:journals/corr/abs-1901-11196}. EDA primarily expands the dataset by four operations: synonym replacement, random insertion, random swap, and random deletion, according to a pre-defined ratio. Moreover, taking inspiration from DAPT~\cite{gururangan2020don}, datasets such as~\cite{kiseleva2022iglu,narayan-chen-etal-2019-collaborative,shi2022learning,zholus2022iglu} are used for performing domain-adaptive fine-tuning. To prevent overfitting, the Fast Gradient Method (FGM) is proposed, inspired by adversarial training, to mitigate the overfitting problem~\cite{goodfellow2014explaining}. Finally, taking cues from~\cite{DBLP:journals/corr/abs-2101-08751}, the list-wise loss is used to train the model.

\section{Conclusions}
\label{sec:conclusions}
In conclusion, our paper addresses the important capability of human intelligence to swiftly adapt to new tasks and multi-modal environments through imitating others and following natural language instructions. We have explored the field of interactive grounded language understanding, aiming to enable seamless human-centered AI collaboration in solving tasks. Specifically, our investigation focuses on the production of clarifying questions in the context of human-AI instruction-based interaction using the Minecraft environment.
By tackling this challenge, we contribute to the development of embodied agents that can effectively understand and respond to human language instructions, as well as ask relevant clarifying questions when necessary. Our work emphasizes the importance of bridging the gap between human communication and AI systems, with the ultimate goal of enhancing user experience and achieving more user-friendly human-AI interactions.

One significant obstacle hindering progress in this field has been the scarcity of appropriate datasets and scalable data collection tools. To address this challenge, we have developed a crowdsourcing tool specifically designed for collecting interactive grounded language instructions within a Minecraft-like environment at a large scale. Additionally, we have created the largest available dataset of human-to-human grounded language instructions, accompanied by clarifying questions. This dataset serves as a valuable resource for various research directions.
Furthermore, we have established baselines for predicting clarifying questions, providing a benchmark for evaluating the performance of future models and algorithms in this domain.

Our contributions lay a solid foundation for further advancements in grounded language understanding research and open up new avenues for exploration and innovation in the field. We believe that our work will inspire and empower researchers to delve deeper into the realm of human-AI interactions, ultimately leading to more effective and seamless collaboration between humans and intelligent embodied agents.

\bibliography{acl}

\begin{thebibliography}{76}
\expandafter\ifx\csname natexlab\endcsname\relax\def\natexlab#1{#1}\fi

\bibitem[{Abramson et~al.(2020)Abramson, Ahuja, Barr, Brussee, Carnevale,
  Cassin, Chhaparia, Clark, Damoc, Dudzik et~al.}]{abramson2020imitating}
Josh Abramson, Arun Ahuja, Iain Barr, Arthur Brussee, Federico Carnevale, Mary
  Cassin, Rachita Chhaparia, Stephen Clark, Bogdan Damoc, Andrew Dudzik, et~al.
  2020.
\newblock Imitating interactive intelligence.
\newblock \emph{arXiv preprint arXiv:2012.05672}.

\bibitem[{Adiwardana et~al.(2020)Adiwardana, Luong, So, Hall, Fiedel,
  Thoppilan, Yang, Kulshreshtha, Nemade, Lu et~al.}]{adiwardana2020towards}
Daniel Adiwardana, Minh-Thang Luong, David~R So, Jamie Hall, Noah Fiedel, Romal
  Thoppilan, Zi~Yang, Apoorv Kulshreshtha, Gaurav Nemade, Yifeng Lu, et~al.
  2020.
\newblock Towards a human-like open-domain chatbot.
\newblock \emph{arXiv preprint arXiv:2001.09977}.

\bibitem[{Aliannejadi et~al.(2020{\natexlab{a}})Aliannejadi, Kiseleva, Chuklin,
  Dalton, and Burtsev}]{Aliannejadi_convAI3}
Mohammad Aliannejadi, Julia Kiseleva, Aleksandr Chuklin, Jeff Dalton, and
  Mikhail Burtsev. 2020{\natexlab{a}}.
\newblock \href {http://arxiv.org/abs/2009.11352} {Convai3: Generating
  clarifying questions for open-domain dialogue systems (clariq)}.

\bibitem[{Aliannejadi et~al.(2020{\natexlab{b}})Aliannejadi, Kiseleva, Chuklin,
  Dalton, and Burtsev}]{aliannejadi2020convai3}
Mohammad Aliannejadi, Julia Kiseleva, Aleksandr Chuklin, Jeff Dalton, and
  Mikhail Burtsev. 2020{\natexlab{b}}.
\newblock Convai3: Generating clarifying questions for open-domain dialogue
  systems (clariq).
\newblock \emph{arXiv preprint arXiv:2009.11352}.

\bibitem[{Aliannejadi et~al.(2021{\natexlab{a}})Aliannejadi, Kiseleva, Chuklin,
  Dalton, and Burtsev}]{aliannejadi-etal-2021-building}
Mohammad Aliannejadi, Julia Kiseleva, Aleksandr Chuklin, Jeff Dalton, and
  Mikhail Burtsev. 2021{\natexlab{a}}.
\newblock \href {https://doi.org/10.18653/v1/2021.emnlp-main.367} {Building and
  evaluating open-domain dialogue corpora with clarifying questions}.
\newblock In \emph{Proceedings of the 2021 Conference on Empirical Methods in
  Natural Language Processing}, pages 4473--4484, Online and Punta Cana,
  Dominican Republic. Association for Computational Linguistics.

\bibitem[{Aliannejadi et~al.(2021{\natexlab{b}})Aliannejadi, Kiseleva, Chuklin,
  Dalton, and Burtsev}]{aliannejadi2021building}
Mohammad Aliannejadi, Julia Kiseleva, Aleksandr Chuklin, Jeffrey Dalton, and
  Mikhail Burtsev. 2021{\natexlab{b}}.
\newblock Building and evaluating open-domain dialogue corpora with clarifying
  questions.
\newblock \emph{arXiv preprint arXiv:2109.05794}.

\bibitem[{Aliannejadi et~al.(2019)Aliannejadi, Zamani, Crestani, and
  Croft}]{aliannejadi2019asking}
Mohammad Aliannejadi, Hamed Zamani, Fabio Crestani, and W~Bruce Croft. 2019.
\newblock Asking clarifying questions in open-domain information-seeking
  conversations.
\newblock In \emph{Proceedings of the 42nd international acm sigir conference
  on research and development in information retrieval}, pages 475--484.

\bibitem[{Arabzadeh et~al.(2022)Arabzadeh, Seifikar, and
  Clarke}]{arabzadeh2022unsupervised}
Negar Arabzadeh, Mahsa Seifikar, and Charles~LA Clarke. 2022.
\newblock Unsupervised question clarity prediction through retrieved item
  coherency.
\newblock In \emph{Proceedings of the 31st ACM International Conference on
  Information \& Knowledge Management}, pages 3811--3816.

\bibitem[{Bara et~al.(2021)Bara, CH-Wang, and Chai}]{bara2021mindcraft}
Cristian-Paul Bara, Sky CH-Wang, and Joyce Chai. 2021.
\newblock Mindcraft: Theory of mind modeling for situated dialogue in
  collaborative tasks.
\newblock \emph{arXiv preprint arXiv:2109.06275}.

\bibitem[{Berant et~al.(2013)Berant, Chou, Frostig, and
  Liang}]{berant2013semantic}
Jonathan Berant, Andrew Chou, Roy Frostig, and Percy Liang. 2013.
\newblock Semantic parsing on freebase from question-answer pairs.
\newblock In \emph{Proceedings of the 2013 conference on empirical methods in
  natural language processing}, pages 1533--1544.

\bibitem[{Bisk et~al.(2016)Bisk, Yuret, and Marcu}]{bisk2016natural}
Yonatan Bisk, Deniz Yuret, and Daniel Marcu. 2016.
\newblock Natural language communication with robots.
\newblock In \emph{Proceedings of the 2016 Conference of the North American
  Chapter of the Association for Computational Linguistics: Human Language
  Technologies}, pages 751--761.

\bibitem[{Brown et~al.(2020)Brown, Mann, Ryder, Subbiah, Kaplan, Dhariwal,
  Neelakantan, Shyam, Sastry, Askell, Agarwal, Herbert-Voss, Krueger, Henighan,
  Child, Ramesh, Ziegler, Wu, Winter, Hesse, Chen, Sigler, Litwin, Gray, Chess,
  Clark, Berner, McCandlish, Radford, Sutskever, and
  Amodei}]{brown2020language}
Tom~B. Brown, Benjamin Mann, Nick Ryder, Melanie Subbiah, Jared Kaplan,
  Prafulla Dhariwal, Arvind Neelakantan, Pranav Shyam, Girish Sastry, Amanda
  Askell, Sandhini Agarwal, Ariel Herbert-Voss, Gretchen Krueger, Tom Henighan,
  Rewon Child, Aditya Ramesh, Daniel~M. Ziegler, Jeffrey Wu, Clemens Winter,
  Christopher Hesse, Mark Chen, Eric Sigler, Mateusz Litwin, Scott Gray,
  Benjamin Chess, Jack Clark, Christopher Berner, Sam McCandlish, Alec Radford,
  Ilya Sutskever, and Dario Amodei. 2020.
\newblock \href {http://arxiv.org/abs/2005.14165} {Language models are few-shot
  learners}.

\bibitem[{Burtsev et~al.(2017)Burtsev, Chuklin, Kiseleva, and
  Borisov}]{burtsev2017search}
Mikhail Burtsev, Aleksandr Chuklin, Julia Kiseleva, and Alexey Borisov. 2017.
\newblock Search-oriented conversational ai (scai).
\newblock In \emph{Proceedings of the ACM SIGIR International Conference on
  Theory of Information Retrieval}, pages 333--334.

\bibitem[{Carta et~al.(2023)Carta, Romac, Wolf, Lamprier, Sigaud, and
  Oudeyer}]{carta2023grounding}
Thomas Carta, Cl{\'e}ment Romac, Thomas Wolf, Sylvain Lamprier, Olivier Sigaud,
  and Pierre-Yves Oudeyer. 2023.
\newblock Grounding large language models in interactive environments with
  online reinforcement learning.
\newblock \emph{arXiv preprint arXiv:2302.02662}.

\bibitem[{Chowdhery et~al.(2022)Chowdhery, Narang, Devlin, Bosma, Mishra,
  Roberts, Barham, Chung, Sutton, Gehrmann et~al.}]{chowdhery2022palm}
Aakanksha Chowdhery, Sharan Narang, Jacob Devlin, Maarten Bosma, Gaurav Mishra,
  Adam Roberts, Paul Barham, Hyung~Won Chung, Charles Sutton, Sebastian
  Gehrmann, et~al. 2022.
\newblock Palm: Scaling language modeling with pathways.
\newblock \emph{arXiv preprint arXiv:2204.02311}.

\bibitem[{Clark et~al.(2020)Clark, Luong, Le, and Manning}]{clark2020electra}
Kevin Clark, Minh-Thang Luong, Quoc~V Le, and Christopher~D Manning. 2020.
\newblock Electra: Pre-training text encoders as discriminators rather than
  generators.
\newblock \emph{arXiv preprint arXiv:2003.10555}.

\bibitem[{Codd(1974)}]{codd1974seven}
Edgar~F Codd. 1974.
\newblock \emph{Seven steps to rendezvous with the casual user}.
\newblock IBM Corporation.

\bibitem[{Copestake and Jones(1990)}]{copestake1990natural}
Ann Copestake and Karen~Sparck Jones. 1990.
\newblock Natural language interfaces to databases.

\bibitem[{Desai et~al.(2016)Desai, Gulwani, Hingorani, Jain, Karkare, Marron,
  Roy et~al.}]{desai2016program}
Aditya Desai, Sumit Gulwani, Vineet Hingorani, Nidhi Jain, Amey Karkare, Mark
  Marron, Subhajit Roy, et~al. 2016.
\newblock Program synthesis using natural language.
\newblock In \emph{Proceedings of the 38th International Conference on Software
  Engineering}, pages 345--356. ACM.

\bibitem[{Devlin et~al.(2018)Devlin, Chang, Lee, and
  Toutanova}]{devlin2018bert}
Jacob Devlin, Ming-Wei Chang, Kenton Lee, and Kristina Toutanova. 2018.
\newblock Bert: Pre-training of deep bidirectional transformers for language
  understanding.
\newblock In \emph{Conference of the North American Chapter of the Association
  for Computational Linguistics: Human Language Technologies (NAACL)}.

\bibitem[{Dinan et~al.(2020)Dinan, Logacheva, Malykh, Miller, Shuster, Urbanek,
  Kiela, Szlam, Serban, Lowe et~al.}]{dinan2020second}
Emily Dinan, Varvara Logacheva, Valentin Malykh, Alexander Miller, Kurt
  Shuster, Jack Urbanek, Douwe Kiela, Arthur Szlam, Iulian Serban, Ryan Lowe,
  et~al. 2020.
\newblock The second conversational intelligence challenge (convai2).
\newblock In \emph{The NeurIPS'18 Competition}, pages 187--208. Springer, Cham.

\bibitem[{Elgohary et~al.(2020)Elgohary, Hosseini, and
  Hassan~Awadallah}]{elgohary-etal-2020-speak}
Ahmed Elgohary, Saghar Hosseini, and Ahmed Hassan~Awadallah. 2020.
\newblock \href {https://doi.org/10.18653/v1/2020.acl-main.187} {Speak to your
  parser: Interactive text-to-{SQL} with natural language feedback}.
\newblock In \emph{Proceedings of the 58th Annual Meeting of the Association
  for Computational Linguistics}, pages 2065--2077, Online. Association for
  Computational Linguistics.

\bibitem[{Fast et~al.(2018)Fast, Chen, Mendelsohn, Bassen, and
  Bernstein}]{fast2018iris}
Ethan Fast, Binbin Chen, Julia Mendelsohn, Jonathan Bassen, and Michael~S
  Bernstein. 2018.
\newblock Iris: A conversational agent for complex tasks.
\newblock In \emph{Proceedings of the 2018 CHI Conference on Human Factors in
  Computing Systems}, page 473. ACM.

\bibitem[{Gao et~al.(2021)Gao, Dai, and
  Callan}]{DBLP:journals/corr/abs-2101-08751}
Luyu Gao, Zhuyun Dai, and Jamie Callan. 2021.
\newblock \href {http://arxiv.org/abs/2101.08751} {Rethink training of {BERT}
  rerankers in multi-stage retrieval pipeline}.
\newblock \emph{CoRR}, abs/2101.08751.

\bibitem[{Gluck and Laird(2018)}]{gluck2018interactive}
Kevin~A Gluck and John~E Laird. 2018.
\newblock \emph{Interactive task learning: Humans, robots, and agents acquiring
  new tasks through natural interactions.}
\newblock The MIT Press.

\bibitem[{Goodfellow et~al.(2014)Goodfellow, Shlens, and
  Szegedy}]{goodfellow2014explaining}
Ian~J Goodfellow, Jonathon Shlens, and Christian Szegedy. 2014.
\newblock Explaining and harnessing adversarial examples.
\newblock \emph{arXiv preprint arXiv:1412.6572}.

\bibitem[{Gray et~al.(2019{\natexlab{a}})Gray, Srinet, Jernite, Yu, Chen, Guo,
  Goyal, Zitnick, and Szlam}]{gray2019craftassist}
Jonathan Gray, Kavya Srinet, Yacine Jernite, Haonan Yu, Zhuoyuan Chen, Demi
  Guo, Siddharth Goyal, C.~Lawrence Zitnick, and Arthur Szlam.
  2019{\natexlab{a}}.
\newblock \href {http://arxiv.org/abs/1907.08584} {Craftassist: A framework for
  dialogue-enabled interactive agents}.

\bibitem[{Gray et~al.(2019{\natexlab{b}})Gray, Srinet, Jernite, Yu, Chen, Guo,
  Goyal, Zitnick, and Szlam}]{gray_craftassist_2019}
Jonathan Gray, Kavya Srinet, Yacine Jernite, Haonan Yu, Zhuoyuan Chen, Demi
  Guo, Siddharth Goyal, C.~Lawrence Zitnick, and Arthur Szlam.
  2019{\natexlab{b}}.
\newblock \href {http://arxiv.org/abs/1907.08584} {{CraftAssist}: {A}
  {Framework} for {Dialogue}-enabled {Interactive} {Agents}}.
\newblock \emph{arXiv:1907.08584 [cs]}.
\newblock ArXiv: 1907.08584.

\bibitem[{Gururangan et~al.(2020)Gururangan, Marasovi{\'c}, Swayamdipta, Lo,
  Beltagy, Downey, and Smith}]{gururangan2020don}
Suchin Gururangan, Ana Marasovi{\'c}, Swabha Swayamdipta, Kyle Lo, Iz~Beltagy,
  Doug Downey, and Noah~A Smith. 2020.
\newblock Don't stop pretraining: Adapt language models to domains and tasks.
\newblock \emph{arXiv preprint arXiv:2004.10964}.

\bibitem[{Hendrix et~al.(1978)Hendrix, Sacerdoti, Sagalowicz, and
  Slocum}]{hendrix1978developing}
Gary~G Hendrix, Earl~D Sacerdoti, Daniel Sagalowicz, and Jonathan Slocum. 1978.
\newblock Developing a natural language interface to complex data.
\newblock \emph{ACM Transactions on Database Systems (TODS)}, 3(2):105--147.

\bibitem[{Izacard et~al.(2021)Izacard, Caron, Hosseini, Riedel, Bojanowski,
  Joulin, and Grave}]{izacard2021towards}
Gautier Izacard, Mathilde Caron, Lucas Hosseini, Sebastian Riedel, Piotr
  Bojanowski, Armand Joulin, and Edouard Grave. 2021.
\newblock Towards unsupervised dense information retrieval with contrastive
  learning.
\newblock \emph{arXiv preprint arXiv:2112.09118}.

\bibitem[{Jayannavar et~al.(2020)Jayannavar, Narayan-Chen, and
  Hockenmaier}]{jayannavar2020learning}
Prashant Jayannavar, Anjali Narayan-Chen, and Julia Hockenmaier. 2020.
\newblock Learning to execute instructions in a minecraft dialogue.
\newblock In \emph{Proceedings of the 58th annual meeting of the association
  for computational linguistics}, pages 2589--2602.

\bibitem[{Johnson et~al.(2016)Johnson, Hofmann, Hutton, and
  Bignell}]{johnson2016malmo}
Matthew Johnson, Katja Hofmann, Tim Hutton, and David Bignell. 2016.
\newblock The malmo platform for artificial intelligence experimentation.
\newblock In \emph{IJCAI}, pages 4246--4247. Citeseer.

\bibitem[{Karpukhin et~al.(2020)Karpukhin, O{\u{g}}uz, Min, Lewis, Wu, Edunov,
  Chen, and Yih}]{karpukhin2020dense}
Vladimir Karpukhin, Barlas O{\u{g}}uz, Sewon Min, Patrick Lewis, Ledell Wu,
  Sergey Edunov, Danqi Chen, and Wen-tau Yih. 2020.
\newblock Dense passage retrieval for open-domain question answering.
\newblock \emph{arXiv preprint arXiv:2004.04906}.

\bibitem[{Kiseleva et~al.(2022{\natexlab{a}})Kiseleva, Li, Aliannejadi,
  Mohanty, ter Hoeve, Burtsev, Skrynnik, Zholus, Panov, Srinet, Szlam, Sun,
  Hofmann, Côté, Awadallah, Abdrazakov, Churin, Manggala, Naszadi, van~der
  Meer, and Kim}]{kiseleva2022interactive}
Julia Kiseleva, Ziming Li, Mohammad Aliannejadi, Shrestha Mohanty, Maartje ter
  Hoeve, Mikhail Burtsev, Alexey Skrynnik, Artem Zholus, Aleksandr Panov, Kavya
  Srinet, Arthur Szlam, Yuxuan Sun, Katja Hofmann, Marc-Alexandre Côté, Ahmed
  Awadallah, Linar Abdrazakov, Igor Churin, Putra Manggala, Kata Naszadi,
  Michiel van~der Meer, and Taewoon Kim. 2022{\natexlab{a}}.
\newblock Interactive grounded language understanding in a collaborative
  environment: Iglu 2021.
\newblock In \emph{NeurIPS 2021 Competitions and Demonstrations Track}, pages
  146--161. PMLR.

\bibitem[{Kiseleva et~al.(2021)Kiseleva, Li, Aliannejadi, Mohanty, ter Hoeve,
  Burtsev, Skrynnik, Zholus, Panov, Srinet et~al.}]{kiseleva2021neurips}
Julia Kiseleva, Ziming Li, Mohammad Aliannejadi, Shrestha Mohanty, Maartje ter
  Hoeve, Mikhail Burtsev, Alexey Skrynnik, Artem Zholus, Aleksandr Panov, Kavya
  Srinet, et~al. 2021.
\newblock Neurips 2021 competition iglu: Interactive grounded language
  understanding in a collaborative environment.
\newblock \emph{arXiv preprint arXiv:2110.06536}.

\bibitem[{Kiseleva et~al.(2022{\natexlab{b}})Kiseleva, Skrynnik, Zholus,
  Mohanty, Arabzadeh, C{\^o}t{\'e}, Aliannejadi, Teruel, Li, Burtsev
  et~al.}]{kiseleva2022iglu}
Julia Kiseleva, Alexey Skrynnik, Artem Zholus, Shrestha Mohanty, Negar
  Arabzadeh, Marc-Alexandre C{\^o}t{\'e}, Mohammad Aliannejadi, Milagro Teruel,
  Ziming Li, Mikhail Burtsev, et~al. 2022{\natexlab{b}}.
\newblock Iglu 2022: Interactive grounded language understanding in a
  collaborative environment at neurips 2022.
\newblock \emph{arXiv preprint arXiv:2205.13771}.

\bibitem[{Kiseleva et~al.(2016{\natexlab{a}})Kiseleva, Williams,
  Hassan~Awadallah, Crook, Zitouni, and Anastasakos}]{kiseleva2016predicting}
Julia Kiseleva, Kyle Williams, Ahmed Hassan~Awadallah, Aidan~C Crook, Imed
  Zitouni, and Tasos Anastasakos. 2016{\natexlab{a}}.
\newblock Predicting user satisfaction with intelligent assistants.
\newblock In \emph{Proceedings of the 39th International ACM SIGIR conference
  on Research and Development in Information Retrieval}, pages 45--54.

\bibitem[{Kiseleva et~al.(2016{\natexlab{b}})Kiseleva, Williams, Jiang,
  Hassan~Awadallah, Crook, Zitouni, and
  Anastasakos}]{kiseleva2016understanding}
Julia Kiseleva, Kyle Williams, Jiepu Jiang, Ahmed Hassan~Awadallah, Aidan~C
  Crook, Imed Zitouni, and Tasos Anastasakos. 2016{\natexlab{b}}.
\newblock Understanding user satisfaction with intelligent assistants.
\newblock In \emph{Proceedings of the 2016 ACM on Conference on Human
  Information Interaction and Retrieval}, pages 121--130.

\bibitem[{K{\"o}hn et~al.(2020)K{\"o}hn, Wichlacz, Sch{\"a}fer, Torralba,
  Hoffmann, and Koller}]{kohn2020mc}
Arne K{\"o}hn, Julia Wichlacz, Christine Sch{\"a}fer, Alvaro Torralba, J{\"o}rg
  Hoffmann, and Alexander Koller. 2020.
\newblock Mc-saar-instruct: a platform for minecraft instruction giving agents.
\newblock In \emph{Proceedings of the 21th Annual Meeting of the Special
  Interest Group on Discourse and Dialogue}, pages 53--56.

\bibitem[{Li et~al.(2020{\natexlab{a}})Li, Mitchell, and
  Myers}]{li-etal-2020-interactive}
Toby Jia-Jun Li, Tom Mitchell, and Brad Myers. 2020{\natexlab{a}}.
\newblock Interactive task learning from {GUI}-grounded natural language
  instructions and demonstrations.
\newblock In \emph{Proceedings of the 58th Annual Meeting of the Association
  for Computational Linguistics: System Demonstrations}.

\bibitem[{Li et~al.(2019)Li, Kiseleva, and De~Rijke}]{li2019dialogue}
Ziming Li, Julia Kiseleva, and Maarten De~Rijke. 2019.
\newblock Dialogue generation: From imitation learning to inverse reinforcement
  learning.
\newblock In \emph{Proceedings of the AAAI conference on artificial
  intelligence}, volume~33, pages 6722--6729.

\bibitem[{Li et~al.(2020{\natexlab{b}})Li, Kiseleva, and
  de~Rijke}]{li-etal-2020-rethinking}
Ziming Li, Julia Kiseleva, and Maarten de~Rijke. 2020{\natexlab{b}}.
\newblock \href {https://doi.org/10.18653/v1/2020.findings-emnlp.316}
  {Rethinking supervised learning and reinforcement learning in task-oriented
  dialogue systems}.
\newblock In \emph{Findings of the Association for Computational Linguistics:
  EMNLP 2020}, pages 3537--3546, Online. Association for Computational
  Linguistics.

\bibitem[{Li et~al.(2021)Li, Kiseleva, and de~Rijke}]{li2021improving}
Ziming Li, Julia Kiseleva, and Maarten de~Rijke. 2021.
\newblock Improving response quality with backward reasoning in open-domain
  dialogue systems.
\newblock In \emph{Proceedings of the 44th International ACM SIGIR Conference
  on Research and Development in Information Retrieval}, pages 1940--1944.

\bibitem[{Li et~al.(2020{\natexlab{c}})Li, Lee, Peng, Li, Kiseleva, de~Rijke,
  Shayandeh, and Gao}]{li2020guided}
Ziming Li, Sungjin Lee, Baolin Peng, Jinchao Li, Julia Kiseleva, Maarten
  de~Rijke, Shahin Shayandeh, and Jianfeng Gao. 2020{\natexlab{c}}.
\newblock \href {https://doi.org/10.18653/v1/2020.findings-emnlp.209} {Guided
  dialogue policy learning without adversarial learning in the loop}.
\newblock In \emph{Findings of the Association for Computational Linguistics:
  EMNLP 2020}, pages 2308--2317, Online. Association for Computational
  Linguistics.

\bibitem[{Liu and Lane(2017)}]{liu2017iterative}
Bing Liu and Ian Lane. 2017.
\newblock Iterative policy learning in end-to-end trainable task-oriented
  neural dialog models.
\newblock In \emph{2017 IEEE Automatic Speech Recognition and Understanding
  Workshop (ASRU)}, pages 482--489. IEEE.

\bibitem[{Liu and Lane(2018)}]{liu2018adversarial}
Bing Liu and Ian Lane. 2018.
\newblock Adversarial learning of task-oriented neural dialog models.
\newblock In \emph{Proceedings of the SIGDIAL 2018 Conference}, pages 350--359.

\bibitem[{Liu et~al.(2019)Liu, Ott, Goyal, Du, Joshi, Chen, Levy, Lewis,
  Zettlemoyer, and Stoyanov}]{LiuRoberta_2019}
Yinhan Liu, Myle Ott, Naman Goyal, Jingfei Du, Mandar Joshi, Danqi Chen, Omer
  Levy, Mike Lewis, Luke Zettlemoyer, and Veselin Stoyanov. 2019.
\newblock \href {http://arxiv.org/abs/1907.11692} {Roberta: {A} robustly
  optimized {BERT} pretraining approach}.
\newblock \emph{CoRR}, abs/1907.11692.

\bibitem[{Mehta et~al.(2023)Mehta, Teruel, Sanz, Deng, Awadallah, and
  Kiseleva}]{mehta2023improving}
Nikhil Mehta, Milagro Teruel, Patricio~Figueroa Sanz, Xin Deng, Ahmed~Hassan
  Awadallah, and Julia Kiseleva. 2023.
\newblock Improving grounded language understanding in a collaborative
  environment by interacting with agents through help feedback.
\newblock \emph{arXiv preprint arXiv:2304.10750}.

\bibitem[{Mohanty et~al.(2022)Mohanty, Arabzadeh, Teruel, Sun, Zholus,
  Skrynnik, Burtsev, Srinet, Panov, Szlam, Côté, and
  Kiseleva}]{mohanty2022collecting}
Shrestha Mohanty, Negar Arabzadeh, Milagro Teruel, Yuxuan Sun, Artem Zholus,
  Alexey Skrynnik, Mikhail Burtsev, Kavya Srinet, Aleksandr Panov, Arthur
  Szlam, Marc-Alexandre Côté, and Julia Kiseleva. 2022.
\newblock Collecting interactive multi-modal datasets for grounded language
  understanding.
\newblock \emph{arXiv preprint arXiv:2211.06552}.

\bibitem[{Narayan-Chen et~al.(2019)Narayan-Chen, Jayannavar, and
  Hockenmaier}]{narayan-chen-etal-2019-collaborative}
Anjali Narayan-Chen, Prashant Jayannavar, and Julia Hockenmaier. 2019.
\newblock \href {https://doi.org/10.18653/v1/P19-1537} {Collaborative dialogue
  in {M}inecraft}.
\newblock In \emph{Proceedings of the 57th Annual Meeting of the Association
  for Computational Linguistics}, pages 5405--5415, Florence, Italy.
  Association for Computational Linguistics.

\bibitem[{Nguyen et~al.(2016)Nguyen, Rosenberg, Song, Gao, Tiwary, Majumder,
  and Deng}]{nguyen2016ms}
Tri Nguyen, Mir Rosenberg, Xia Song, Jianfeng Gao, Saurabh Tiwary, Rangan
  Majumder, and Li~Deng. 2016.
\newblock Ms marco: A human generated machine reading comprehension dataset.
\newblock \emph{choice}, 2640:660.

\bibitem[{Ogawa et~al.(2020)Ogawa, Nishikawa, Tokunaga, and
  Yokono}]{ogawa-etal-2020-gamification}
Haruna Ogawa, Hitoshi Nishikawa, Takenobu Tokunaga, and Hikaru Yokono. 2020.
\newblock \href {https://www.aclweb.org/anthology/2020.lrec-1.876}
  {Gamification platform for collecting task-oriented dialogue data}.
\newblock In \emph{Proceedings of the 12th Language Resources and Evaluation
  Conference}, pages 7084--7093, Marseille, France. European Language Resources
  Association.

\bibitem[{OpenAI(2023)}]{2303.08774}
OpenAI. 2023.
\newblock \href {http://arxiv.org/abs/arXiv:2303.08774} {Gpt-4 technical
  report}.

\bibitem[{Park et~al.(2023)Park, O'Brien, Cai, Morris, Liang, and
  Bernstein}]{park2023generative}
Joon~Sung Park, Joseph~C O'Brien, Carrie~J Cai, Meredith~Ringel Morris, Percy
  Liang, and Michael~S Bernstein. 2023.
\newblock Generative agents: Interactive simulacra of human behavior.
\newblock \emph{arXiv preprint arXiv:2304.03442}.

\bibitem[{Press et~al.(2022)Press, Zhang, Min, Schmidt, Smith, and
  Lewis}]{press2022measuring}
Ofir Press, Muru Zhang, Sewon Min, Ludwig Schmidt, Noah~A Smith, and Mike
  Lewis. 2022.
\newblock Measuring and narrowing the compositionality gap in language models.
\newblock \emph{arXiv preprint arXiv:2210.03350}.

\bibitem[{Reimers and Gurevych(2019)}]{reimers2019sentence}
Nils Reimers and Iryna Gurevych. 2019.
\newblock Sentence-bert: Sentence embeddings using siamese bert-networks.
\newblock \emph{arXiv preprint arXiv:1908.10084}.

\bibitem[{Roller et~al.(2020)Roller, Dinan, Goyal, Ju, Williamson, Liu, Xu,
  Ott, Shuster, Smith et~al.}]{roller2020recipes}
Stephen Roller, Emily Dinan, Naman Goyal, Da~Ju, Mary Williamson, Yinhan Liu,
  Jing Xu, Myle Ott, Kurt Shuster, Eric~M Smith, et~al. 2020.
\newblock Recipes for building an open-domain chatbot.
\newblock \emph{arXiv preprint arXiv:2004.13637}.

\bibitem[{Shi et~al.(2022)Shi, Feng, and Lipani}]{shi2022learning}
Zhengxiang Shi, Yue Feng, and Aldo Lipani. 2022.
\newblock Learning to execute or ask clarification questions.
\newblock \emph{arXiv preprint arXiv:2204.08373}.

\bibitem[{Shi et~al.(2023)Shi, Ramos, Kim, Wang, Rahmani, and
  Lipani}]{shi2023ask}
Zhengxiang Shi, Jerome Ramos, To~Eun Kim, Xi~Wang, Hossein~A. Rahmani, and Aldo
  Lipani. 2023.
\newblock \href {http://arxiv.org/abs/2305.05754} {When and what to ask through
  world states and text instructions: Iglu nlp challenge solution}.

\bibitem[{Shridhar et~al.(2020)Shridhar, Thomason, Gordon, Bisk, Han, Mottaghi,
  Zettlemoyer, and Fox}]{shridhar2020alfred}
Mohit Shridhar, Jesse Thomason, Daniel Gordon, Yonatan Bisk, Winson Han,
  Roozbeh Mottaghi, Luke Zettlemoyer, and Dieter Fox. 2020.
\newblock Alfred: A benchmark for interpreting grounded instructions for
  everyday tasks.
\newblock In \emph{Proceedings of the IEEE/CVF conference on computer vision
  and pattern recognition}, pages 10740--10749.

\bibitem[{Skrynnik et~al.(2022)Skrynnik, Volovikova, C{\^o}t{\'e}, Voronov,
  Zholus, Arabzadeh, Mohanty, Teruel, Awadallah, Panov, Burtsev, and
  Kiseleva}]{skrynnik2022learning}
Alexey Skrynnik, Zoya Volovikova, Marc-Alexandre C{\^o}t{\'e}, Anton Voronov,
  Artem Zholus, Negar Arabzadeh, Shrestha Mohanty, Milagro Teruel, Ahmed
  Awadallah, Aleksandr Panov, Mikhail Burtsev, and Julia Kiseleva. 2022.
\newblock Learning to solve voxel building embodied tasks from pixels and
  natural language instructions.
\newblock \emph{arXiv preprint arXiv:2211.00688}.

\bibitem[{Srinet et~al.(2020)Srinet, Jernite, Gray, and
  Szlam}]{srinet-etal-2020-craftassist}
Kavya Srinet, Yacine Jernite, Jonathan Gray, and Arthur Szlam. 2020.
\newblock \href {https://doi.org/10.18653/v1/2020.acl-main.427}
  {{C}raft{A}ssist instruction parsing: Semantic parsing for a voxel-world
  assistant}.
\newblock In \emph{Proceedings of the 58th Annual Meeting of the Association
  for Computational Linguistics}, pages 4693--4714, Online. Association for
  Computational Linguistics.

\bibitem[{Su et~al.(2017)Su, Awadallah, Khabsa, Pantel, Gamon, and
  Encarnacion}]{su2017building}
Yu~Su, Ahmed~Hassan Awadallah, Madian Khabsa, Patrick Pantel, Michael Gamon,
  and Mark Encarnacion. 2017.
\newblock Building natural language interfaces to web apis.
\newblock In \emph{Proceedings of the 2017 ACM on Conference on Information and
  Knowledge Management}, pages 177--186. ACM.

\bibitem[{Szlam et~al.(2019)Szlam, Gray, Srinet, Jernite, Joulin, Synnaeve,
  Kiela, Yu, Chen, Goyal, Guo, Rothermel, Zitnick, and Weston}]{szlam_why_2019}
Arthur Szlam, Jonathan Gray, Kavya Srinet, Yacine Jernite, Armand Joulin,
  Gabriel Synnaeve, Douwe Kiela, Haonan Yu, Zhuoyuan Chen, Siddharth Goyal,
  Demi Guo, Danielle Rothermel, C.~Lawrence Zitnick, and Jason Weston. 2019.
\newblock \href {http://arxiv.org/abs/1907.09273} {Why {Build} an {Assistant}
  in {Minecraft}?}
\newblock \emph{arXiv:1907.09273 [cs]}.
\newblock ArXiv: 1907.09273.

\bibitem[{Tellex et~al.(2011)Tellex, Kollar, Dickerson, Walter, Banerjee,
  Teller, and Roy}]{tellex2011understanding}
Stefanie Tellex, Thomas Kollar, Steven Dickerson, Matthew~R Walter, Ashis~Gopal
  Banerjee, Seth Teller, and Nicholas Roy. 2011.
\newblock Understanding natural language commands for robotic navigation and
  mobile manipulation.
\newblock In \emph{Twenty-Fifth AAAI Conference on Artificial Intelligence}.

\bibitem[{Wei and Zou(2019)}]{DBLP:journals/corr/abs-1901-11196}
Jason~W. Wei and Kai Zou. 2019.
\newblock \href {http://arxiv.org/abs/1901.11196} {{EDA:} easy data
  augmentation techniques for boosting performance on text classification
  tasks}.
\newblock \emph{CoRR}, abs/1901.11196.

\bibitem[{Winograd(1972)}]{winograd1972understanding}
Terry Winograd. 1972.
\newblock Understanding natural language.
\newblock \emph{Cognitive psychology}, 3(1):1--191.

\bibitem[{Woods et~al.(1972)Woods, Kaplan, and Webber}]{woods1972lunar}
W.~A. Woods, Ronald~M Kaplan, and Bonnie~L. Webber. 1972.
\newblock The lunar sciences natural language information system: Final report.
\newblock \emph{BBN Report 2378}.

\bibitem[{Yao et~al.(2019)Yao, Su, Sun, and Yih}]{yao-etal-2019-model}
Ziyu Yao, Yu~Su, Huan Sun, and Wen-tau Yih. 2019.
\newblock \href {https://doi.org/10.18653/v1/D19-1547} {Model-based interactive
  semantic parsing: A unified framework and a text-to-{SQL} case study}.
\newblock In \emph{Proceedings of the 2019 Conference on Empirical Methods in
  Natural Language Processing and the 9th International Joint Conference on
  Natural Language Processing (EMNLP-IJCNLP)}, pages 5447--5458, Hong Kong,
  China. Association for Computational Linguistics.

\bibitem[{Yao et~al.(2020)Yao, Tang, Yih, Sun, and Su}]{yao2020imitation}
Ziyu Yao, Yiqi Tang, Wen-tau Yih, Huan Sun, and Yu~Su. 2020.
\newblock An imitation game for learning semantic parsers from user
  interaction.

\bibitem[{Young et~al.(2013)Young, Ga{\v{s}}i{\'c}, Thomson, and
  Williams}]{young2013pomdp}
Steve Young, Milica Ga{\v{s}}i{\'c}, Blaise Thomson, and Jason~D Williams.
  2013.
\newblock Pomdp-based statistical spoken dialog systems: A review.
\newblock \emph{Proceedings of the IEEE}, 101(5):1160--1179.

\bibitem[{Zhan et~al.(2021)Zhan, Mao, Liu, Guo, Zhang, and
  Ma}]{zhan2021optimizing}
Jingtao Zhan, Jiaxin Mao, Yiqun Liu, Jiafeng Guo, Min Zhang, and Shaoping Ma.
  2021.
\newblock Optimizing dense retrieval model training with hard negatives.
\newblock In \emph{Proceedings of the 44th International ACM SIGIR Conference
  on Research and Development in Information Retrieval}, pages 1503--1512.

\bibitem[{Zhang et~al.(2021)Zhang, Jauhar, Kiseleva, White, and
  Roth}]{zhang-etal-2021-learning}
Yi~Zhang, Sujay~Kumar Jauhar, Julia Kiseleva, Ryen White, and Dan Roth. 2021.
\newblock \href {https://doi.org/10.18653/v1/2021.naacl-main.217} {Learning to
  decompose and organize complex tasks}.
\newblock In \emph{Proceedings of the 2021 Conference of the North American
  Chapter of the Association for Computational Linguistics: Human Language
  Technologies}, pages 2726--2735, Online. Association for Computational
  Linguistics.

\bibitem[{Zhang et~al.(2019)Zhang, Sun, Galley, Chen, Brockett, Gao, Gao, Liu,
  and Dolan}]{zhang2019dialogpt}
Yizhe Zhang, Siqi Sun, Michel Galley, Yen-Chun Chen, Chris Brockett, Xiang Gao,
  Jianfeng Gao, Jingjing Liu, and Bill Dolan. 2019.
\newblock Dialogpt: Large-scale generative pre-training for conversational
  response generation.
\newblock \emph{arXiv preprint arXiv:1911.00536}.

\bibitem[{Zholus et~al.(2022)Zholus, Skrynnik, Mohanty, Volovikova, Kiseleva,
  Szlam, Cot{\'e}, and Panov}]{zholus2022iglu}
Artem Zholus, Alexey Skrynnik, Shrestha Mohanty, Zoya Volovikova, Julia
  Kiseleva, Artur Szlam, Marc-Alexandre Cot{\'e}, and Aleksandr~I Panov. 2022.
\newblock Iglu gridworld: Simple and fast environment for embodied dialog
  agents.
\newblock \emph{arXiv preprint arXiv:2206.00142}.

\end{thebibliography}
\bibliographystyle{acl_natbib}

\end{document}